\pdfoutput=1

\documentclass[11pt]{article}

\usepackage{acl}

\usepackage{times}
\usepackage{latexsym}
\usepackage{CJKutf8}

\usepackage[T1]{fontenc}

\usepackage[utf8]{inputenc}
\usepackage{url}
\usepackage{soul}

\usepackage{microtype}
\usepackage{xspace}
\usepackage{booktabs}
\usepackage{makecell}
\usepackage{multirow}
\usepackage{graphicx}
\usepackage{enumitem}
\usepackage[normalem]{ulem}
\usepackage{xpatch}
\xpatchcmd{\sout}
  {\bgroup}
  {\bgroup}
  {}{}

\newcommand{\crnn}{\texttt{c-RNN}\xspace}
\newcommand{\conatt}{\texttt{ConATT}\xspace}
\newcommand{\bert}{\texttt{BERT}\xspace}
\newcommand{\webnlg}{\textsc{w}eb\textsc{nlg}\xspace}
\newcommand{\ontonotes}{\textsc{o}nto\textsc{n}otes\xspace}
\newcommand{\ontonotesen}{\textsc{o}nto\textsc{n}otes-\textsc{en}\xspace}
\newcommand{\ontonoteszh}{\textsc{o}nto\textsc{n}otes-\textsc{zh}\xspace}

\newcommand{\term}[1]{\texttt{#1}}

\title{Assessing Neural Referential Form Selectors on \\a Realistic Multilingual Dataset}

\author{Guanyi Chen\textsuperscript{$\spadesuit$}, 
Fahime Same\textsuperscript{$\heartsuit$}, \and
Kees van Deemter\textsuperscript{$\spadesuit$}\\
\textsuperscript{$\spadesuit$}Department of Information and Computing Sciences, Utrecht University\\
\textsuperscript{$\heartsuit$}Department of Linguistics, University of Cologne\\
\texttt{g.chen@uu.nl, f.same@uni-koeln.de, c.j.vandeemter@uu.nl}}

\begin{document}
\maketitle
\begin{abstract}
Previous work on Neural Referring Expression Generation (REG) all uses \webnlg, an English dataset that has been shown to reflect a very limited range of referring expression (RE) use.
To tackle this issue, we build a dataset based on the \ontonotes corpus that contains a broader range of RE use in both English and Chinese (a language that uses zero pronouns). We build neural Referential Form Selection (RFS) models accordingly, assess them on the dataset and conduct probing experiments. 
The experiments suggest that, compared to \webnlg, \ontonotes is better for assessing REG/RFS models. 
We compare English and Chinese RFS and confirm that in both languages \bert has the highest performance. Also, our results suggest that in line with linguistic theories, Chinese RFS depends more on discourse context than English.
\end{abstract}

\section{Introduction} \label{sec:intro}

\begin{CJK}{UTF8}{gbsn}

Referring Expression Generation (REG) In Context is a key task in the classic Natural Language Generation pipeline~\citep{reiter_dale_2000, 10.5555/3241691.3241693}. Given a discourse whose referring expressions (REs) have yet to be realised and given their intended referents, it aims to develop an algorithm that generates all these REs. 

Traditionally, REG In Context (hereafter REG) is a two-step process. In the first step, the Referential Form (RF) is determined, e.g. whether to use a proper name, a description, a demonstrative or a pronoun. This step is the focus of this work and will be hereafter called Referential Form Selection (RFS). In the second step, the content of the RE is determined. For example, to refer to \emph{Joe Biden}, one needs to choose from options such as ``\emph{the president}'', ``\emph{the 46th president of US}''.

In recent years, many works on REG have started to use neural networks. For example, \citet{castro-ferreira-etal-2018-neuralreg, cao-cheung-2019-referring, cunha-etal-2020-referring} have proposed to generate REs in an End2End manner, i.e., to tackle the selection of form and content simultaneously. \citet{chen-etal-2021-neural-referential} used \texttt{BERT}~\citep{devlin-etal-2019-bert} to perform RFS. One commonality between these studies is that they were all tested on a benchmark dataset, namely \webnlg~\citep{gardent-etal-2017-creating, castro-ferreira-etal-2018-enriching}.



However, \citet{chen-etal-2021-neural-referential} and \citet{same-etal-2022-non} found that \webnlg is not ideal for assessing REG/RFS algorithms because (1) it consists of rather formal texts that may not reflect everyday RE use; (2) its texts are very short and have a simple syntactic structure; and (3) most of its REs are first-mentions. These limitations led to some unexpected results when they tested their RFS models on \webnlg. 
For example, advanced pre-trained models (i.e., \texttt{BERT}) performed worse than simpler models (i.e., single-layer GRU~\citep{cho-etal-2014-properties}) without any pre-training. 
By probing\footnote{Probing is an established method to analyse whether the latent representations of a model encode certain information.} various RFS models, they found that though \texttt{BERT} encodes more linguistic information, which is crucial for RFS, it still performs worse than GRU.
In this study, we are interested in \emph{how well each RFS model performs when tested on a dataset that addresses the above limitations} -- in what follows, we call this a ``realistic" dataset, for short.

Additionally, all the above studies were conducted on English only. It has been pointed out that speakers of East Asian languages (e.g. Chinese and Japanese) use REs differently from speakers of Western European languages (e.g. English and Dutch;~\citet{newnham1971about}). Theoretical linguists~\citep{huang1984distribution} have suggested that East Asian languages rely more heavily on context than Western European languages (see~\citet{chen2022computational} for empirical testing and computational modelling). As a result, speakers of East Asian languages frequently use Zero Pronouns (ZPs), i.e. REs that contain no words and are resolved based merely on their context.\footnote{For example, consider the question in Chinese: ``你看见比尔了吗？'' (\emph{Have you see Bill?}). A Chinese speaker can reply ``$\emptyset$看见$\emptyset$了。'' (\emph{$\emptyset$ saw $\emptyset$.}) where the two $\emptyset$ are ZPs that refer to the speaker himself/herself and ``Bill'' respectively.}
This poses two challenges for REG/RFS models: (1) they need to be better able to encode contextual information; (2) they need to account for an additional RF (i.e. ZP).
Therefore, we are curious to see \emph{how well each RFS model performs when tested on a language that has more RFs and relies more on context than English}.

To answer the research questions above, we construct a ``realistic" multilingual dataset of English and Chinese and try different model architectures, such as models with/without pre-trained word embeddings, and models incorporating \bert. 
We report the results and compare model behaviours on English and Chinese subsets.
The code used in this study is available at: \url{https://github.com/a-quei/probe-neuralreg}.

\end{CJK}
\section{Referential Form Selection (RFS)} \label{sec:task}

\begin{table}[t!]
\small
\centering
\begin{tabular}{p{7cm}}
\toprule
\textbf{Text}: Amatriciana sauce is made with Tomato. It is a traditional Italian sauce. Amatriciana is a sauce containing Tomato that comes from Italy. \\ \midrule
\textbf{Delexicialised Text}:
\hl{Amatriciana\_sauce} is made with Tomato. \hl{Amatriciana\_sauce} is a traditional \hl{Italy} sauce. \hl{Amatriciana\_sauce} is a sauce containing Tomato that comes from \hl{Italy}.\\
\bottomrule
\end{tabular}
\caption{An example data from the \textsc{w}eb\textsc{nlg} corpus. In the delexicalised text, every entity is \\hl{highlighted}.}
\label{tab:webnlg_sample}
\end{table}

Using \webnlg, \citet{castro-ferreira-etal-2018-neuralreg} re-defined the REG task in order to accommodate deep learning techniques. Subsequently, \citet{chen-etal-2021-neural-referential} adapted the definition to fit the RFS task. The first step is to remove from each RE all information about the RF of that RE.
Concretely, as shown in Table~\ref{tab:webnlg_sample}, \citet{castro-ferreira-etal-2018-neuralreg} first ``delexicalised" 
each text in \webnlg by assigning a general entity tag to each entity and replacing all REs referring to that entity with that tag. In most cases, 
a tag is assigned to an entity by replacing whitespaces in its proper name with underscores, e.g. ``\emph{Amatriciana sauce}'' to ``\emph{Amatriciana\_sauce}''.


For a target referent $x^{(r)}$ (e.g. the second ``\emph{Amatriciana\_sauce}'' in Table~\ref{tab:webnlg_sample}), given the referent, its pre-context in the discourse $x^{(pre)}$ (e.g. ``\emph{Amatriciana\_sauce is made with Tomato.}'') and its post-context $x^{(post)}$ (e.g. ``\emph{is a traditional Italy sauce. Amatriciana\_sauce is a sauce containing Tomato that comes from Italy.}''), the RFS task is to decide the proper RF $\hat{f}$ (e.g., pronoun).
\section{Dataset Construction} \label{sec:data}

To construct a realistic multilingual REG/RFS dataset, we used the Chinese and English portions of the \ontonotes dataset\footnote{OntoNotes is licensed under the Linguistic Data Consortium: \url{https://catalog.ldc.upenn.edu/LDC2013T19}.}
whose contents come from six sources, namely broadcast news, newswires, broadcast conversations, telephone conversations, web blogs, and magazines. We call the resulting Chinese subset \ontonoteszh and the English subset \ontonotesen. In the following, we describe the construction process. 

\begin{table}[t!]
\small
\centering
\begin{tabular}{llp{5cm}}
\toprule
\multirow{3}{*}{EN} & 4-Way & Demonstrative, Description, Proper Name, Pronoun \\
& 3-Way & Description, Proper Name, Pronoun \\
& 2-Way & Non-pronominal, Pronominal \\ \midrule
\multirow{4}{*}{ZH} & 5-Way & Demonstrative, Description, Proper Name, Pronoun, ZP \\
& 4-Way & Description, Proper Name, Pronoun, ZP \\
& 3-Way & Non-pronominal, Pronoun, ZP \\
& 2-Way & Overt Referring Expression, ZP \\
\bottomrule
\end{tabular}
\caption{Types of RF classification and possible classes. Demonstratives are grouped with descriptions in 3-way EN and 4-way ZH classifications under the category \textit{Description}. The category \textit{Non-pronominal} contains proper names, descriptions, and demonstratives.}
\label{tab:ontonotes_class}
\end{table}

First, for each RE in \ontonotes, we used the 3 previous sentences as the pre-context and the 3 subsequent sentences as the post-context. 
Similar to \citet{chen-etal-2021-neural-referential}, we are interested in different RF classification tasks. 
For Chinese, for example, we not only have a 2-way classification task where models have to decide whether to use a ZP or an overt RE, but also a 5-way task where models have to choose from a more fine-grained list of possible RFs.
Table~\ref{tab:ontonotes_class} lists all categories in both \ontonotesen and \ontonoteszh.
Using the constituency syntax tree of the sentence containing the target referent and the surface form of the target, we automatically annotated each RE with its RF category. 
For example, an RE is considered a demonstrative if it is annotated in the syntax tree as a noun phrase and its surface form contains a demonstrative determiner.

Second, we excluded all coreferential chains consisting only of pronouns and ZPs. The pronominal chains consist mainly of first/second-person referents, and we do not expect much variation in referential form in these cases. In other words, we only included the chains that have \emph{at least} one overt non-pronominal RE. 

Third, we delexicalised the corpus following \citet{castro-ferreira-etal-2018-neuralreg}. Additionally, since we used the Chinese \bert as one of our RFS models and it only accepts input shorter than 512 characters, 
we removed all samples in \ontonoteszh whose total length (calculated by removing all underscores introduced during delexicalisation and summing the length of pre-contexts, post-contexts, and target referents) is longer than 512 characters.
Experiments with models other than \bert on the original \ontonoteszh show that this does not bias the conclusions of this study (see Appendix~\ref{sec:app_whole}).

Last, we split the whole dataset into a training set and a test set in accordance with the CoNLL 2012 Shared Task~\citep{pradhan-etal-2012-conll}.
Since ZPs in Chinese are only annotated in the training and development sets, following~\citet{chen-ng-2016-chinese}, \citet{chen-etal-2018-modelling}, and~\citet{yin-etal-2018-zero}, we used the development set as the test set and sampled 10\% of the documents from the training set as the development data.
Thus, we obtained \ontonotesen, where the training, development, and test sets contain 71667, 8149, and 7619 samples, respectively, and \ontonoteszh, where the training, development, and test sets contain 70428, 9217, and 11607 samples, respectively.

\paragraph{\ontonotes vs. \webnlg.}

\begin{table}[t!]
\resizebox{\columnwidth}{!}{%
\centering
\begin{tabular}{lccc}
\toprule
 & \webnlg & \textsc{o-en} & \textsc{o-zh} \\ \midrule
 Percentage of First Mentions & 85\% & 43\% & 43\% \\
 Percentage of Proper Names & 71\% & 21\% & 15\% \\
 Average Number of Tokens & 18.62 & 106.44 & 139.55 \\
\bottomrule
\end{tabular}}
\caption{Statistics of \webnlg and \ontonotes. \textsc{o-en} and \textsc{o-zh} stand for \ontonotesen and \ontonoteszh. }
\label{tab:statistics}
\end{table}

Based on the nature of \ontonotes and the statistics in Table~\ref{tab:statistics}, we observe that: (1) the \webnlg data is all from DBPedia, while the \ontonotes data is multi-genre; (2) \ontonotes has a much smaller proportion of first mentions and proper names; and (3) the documents in \ontonotes are on average much longer than those in \webnlg.

Another difference between \webnlg and \ontonotes is in the ratio of seen and unseen entities in their test sets. \citet{castro-ferreira-etal-2018-enriching} divided the documents in the \webnlg's test set into \emph{seen} (where all the data come from the same domains as the training data) and \emph{unseen} (where all the data come from different domains than the training data). Almost all referents from the seen test set appear in the training set (9580 out of 9644), while only a few referents from the unseen test set appear in the training set (688 out of 9644). \footnote{\citet{chen-etal-2021-neural-referential} used only seen entities because the size of the underlying triples of the unseen test set differs from both the training set and seen test set.} In \ontonotes, 38.44\% and 41.45\% of the referents in the test sets of \ontonotesen and \ontonoteszh also appear in the training sets.

Having said this, \ontonotes largely mitigates the problems of \webnlg discussed in \S\ref{sec:intro}. If \ontonotes is a ``better'' and more ``representative" corpus for assessing REG/RFS models, we can expect more ``expected'' results: models with pre-training outperform those without, and models that learn more useful linguistic information outperform those that learn less. We will detail our expectations in \S\ref{sec:hypo}.
\section{Modelling RFS} \label{sec:model}

We introduce how we represent entities and how we adapt the RFS models of \citet{chen-etal-2021-neural-referential}.

\subsection{Entity Representation} \label{sec:entity_rep}

Unlike \webnlg, whose 99.34\% of referents in the test set appear in the training set, the majority of referents in \ontonotes do not appear in both training and test sets. 
This means that RFS models should be able to handle unseen referents, but mapping each entity to a general entity tag with underscores would prevent the models from doing so~\citep{cao-cheung-2019-referring, cunha-etal-2020-referring} because entity tags of unseen entities are usually out-of-vocabulary (OOV) words.
Additionally, when incorporating pre-trained word embeddings and language models, using entity tags prevents entity representations from benefiting from these pre-trained models (again since the entity tags of unseen entities are usually OOV words).

Similar to~\citet{cunha-etal-2020-referring}, we replaced underscores in general entity tags (e.g. ``\emph{Amatriciana\_sauce}'') with whitespaces (henceforth, lexical tags, e.g. ``\emph{Amatriciana sauce}'').
Arguably, there is a trade-off between using entity tags and using lexical tags. In contrast to lexical tags, the use of entity tags helps models identify mentions of the same entity in discourse, which has been shown to be a crucial feature for RFS. However, using entity tags prevents models from dealing with unseen entities and reduces the benefit of using pre-trained language models. In \S\ref{sec:results}, we compare the performance of using entity tags and lexical tags. 

\subsection{RFS Models}

To build the RFS models, we use the two neural models from \citet{chen-etal-2021-neural-referential}: \crnn and \conatt. Given the task definition in \S\ref{sec:task}, models take pre-context $x^{(pre)}$, target referent $x^{(r)}$, and post-context $x^{(post)}$ as inputs.
As a result of using lexical tags, each target referent is no longer a single tag, but a sequence of tokens. In other words, instead of being $\{w_i\}$, $x^{(r)}$ is $\{w_i, w_{i+1}, ..., w_{j}\}$. The other two inputs are pre-context $x^{(pre)} = \{w_1, w_2, ..., w_{i-1}\}$ and post-context $x^{(post)} = \{w_{j+1}, w_{j+1}, ..., w_{n}\}$. The architectures of the models are as follows:
\paragraph{\crnn.} \crnn concatenates $x^{(pre)}$, $x^{(r)}$ and $x^{(post)}$, and uses a single bidirectional GRU to encode them all. Formally, we obtain a sequence of hidden representations by $h = \mbox{BiGRU}([x^{(pre)}, x^{(r)}, x^{(post)}])$.
We then use the summation of the hidden representations at the beginning and the end of the target referent (i.e., $i$ and $j$) for calculating the final representation:
\begin{equation} \label{eq:r_1}
    R = ReLU(W_f [h_i+h_j]),
\end{equation}
where $W_f$ is the weight in the feed-forward layer. 
$R$ is then used for predicting the RF:
\begin{equation} \label{eq:prediction}
    P(\hat{f}|x^{(pre)}, x^{(r)}, x^{(post)}) = \mbox{Softmax}(W_c R),
\end{equation}
where $W_c$ is the weight in the output layer. $x$ can be initialised randomly or initialised by pre-trained word embeddings or language models.
We tested both the vanilla \crnn and \crnn, whose input layer is initialised by pre-trained word embeddings or by \bert. 

\paragraph{\conatt.} \conatt first encodes $x^{(pre)}$, $x^{(r)}$ and $x^{(post)}$ separately using three bidirectional GRUs and three self-attention modules~\citep{yang-etal-2016-hierarchical}. For each input $x^{(k)}$, we first obtain $h^{(k)}$ using a BiGRU: $h^{(k)} = \mbox{BiGRU}(x^{(k)})$.
Subsequently, given the total $M$ steps in $h^{(k)}$, we first calculate the attention weight $\alpha^{(k)}_j$ at each step $j$ by:
\begin{equation}
    \alpha_j^{(k)} = \frac{\exp(e_j^{(k)})}{\sum_{m=1}^M \exp(e_m^{(k)})},
\end{equation}
where $e_j^{(k)} = v_a^{(k)T}\mbox{tanh}(W_a^{(k)}h_j^{(k)})$, $v_a$ is the attention vector and $W_a$ is the weight in the attention layer. The context representation of $x^{(k)}$ is then the weighted sum of $h^{(k)}$: $c^{(k)} = \sum_{j=1}^N \alpha_j^{(k)} h^{(k)}$.

After obtaining $c^{(pre)}$, $c^{(r)}$ and $c^{(post)}$, we concatenate them with the target entity embedding $x^{(r)}$, and pass it through a feed forward network to obtain the final representation:
\begin{equation} \label{eq:r_2}
    R = \mbox{ReLU}(W_f [c^{(pre)}, c^{(r)}, c^{(post)}]),
\end{equation}
where $[\cdot, \cdot]$ represents a concatenation operation.
The prediction is made using Equation~\ref{eq:prediction}.
The input layer of \conatt is initialised either randomly or by pre-trained word embeddings.


\section{Hypotheses} \label{sec:hypo}
\ontonotes reflects a broader range of RE use and is, therefore, more appropriate as a source of insights into the human use of REs.
Thus, it is plausible to expect that the ``unexpected results'' of \S\ref{sec:intro} will not occur when assessing RFS models (see \S\ref{sec:model}) on \ontonotes. 
More specifically, we expect:
\begin{itemize}
    \item[$\mathcal{H}_1$] models that incorporate pre-training (i.e., pre-trained word embeddings and \bert, which has been proved to be effective in many NLP tasks) work better than those that do not;
    \item[$\mathcal{H}_2$] \conatt, which has been shown to perform well on both REG~\citep{castro-ferreira-etal-2018-neuralreg} and co-reference resolution~\citep{yin-etal-2018-zero}, works better than \crnn;
    \item[$\mathcal{H}_3$] models that learn more useful linguistic information (confirmed by probing experiments) perform better than those that learn less. 
\end{itemize}

Comparing Chinese and English, we can see in Table~\ref{tab:ontonotes_class} that Chinese has an additional category compared to English, namely ZP. 
Given the theory that Chinese speakers process ZPs in the same way as pronouns~\citep{yang1999comprehension}, we expect:
\begin{itemize}
    \item[$\mathcal{H}_4$] RFS models that work well in English would also work well in Chinese.
\end{itemize}  
Additionally, since Chinese relies more on context than English (see \S\ref{sec:intro}), it is plausible to expect:
\begin{itemize}
    \item[$\mathcal{H}_5$]  Chinese RFS models would benefit more from the use of contextual representations (i.e., \bert) than English RFS models.
\end{itemize}
\begin{table*}[t!]
\centering
\begin{tabular}{lccccccccc}
\toprule
 & \multicolumn{3}{c}{4-way} & \multicolumn{3}{c}{3-way} & \multicolumn{3}{c}{2-way} \\\cmidrule(lr){2-4} \cmidrule(lr){5-7} \cmidrule(lr){8-10}
Model & P & R & F & P & R & F & P & R & F \\ \midrule
\texttt{XGBoost}  & 48.96  & 49.69  & 49.12 & 67.78  & 65.78  & 66.44  & 79.11  & 78.01 & 78.42  \\ \midrule
\texttt{c-RNN}  & 65.45 & 60.59 & 62.38 & 68.19 & 69.19 & 68.55 & 76.66 & 75.23 & 75.70 \\
\texttt{ +Glove}  & \underline{66.06} & \underline{63.39} & \underline{64.56} & \underline{69.94} & \underline{70.14} & \underline{70.01} & \underline{77.61} & \underline{76.31} & \underline{76.67} \\
\texttt{ +BERT}  & \textbf{73.57} & \textbf{75.94} & \textbf{74.59} & \textbf{80.53} & \textbf{81.81} & \textbf{81.03} & \textbf{87.21} & \textbf{86.97} & \textbf{87.08} \\
& & & (+19.57\%) & & & (+18.21\%) & & & (+15.03\%)\\
\texttt{ConATT}  & 61.29 & 62.21 & 61.58 & 66.34 & 65.87 & 66.01 & 73.19 & 73.21 & 73.19 \\
\texttt{ +Glove}  & 63.71 & 61.70 & 62.51 & 67.18 & 66.88 & 67.00 & 75.17 & 74.48 & 74.75 \\
\bottomrule
\end{tabular}
\caption{Evaluation results of the English RFS systems on \ontonotesen with lexical tags. Best results are \textbf{boldfaced}, whereas the second best results are \underline{underlined}. ``P'', ``R'' and ``F'' stand for macro-averaged precision, recall and F1 score. Each percentage below the F-score of \bert indicates how much \crnn gains from using \bert compared to not using \bert.}
\label{tab:en_cls_result}
\end{table*}
\begin{table*}
\centering
\resizebox{\textwidth}{!}{%
\begin{tabular}{lcccccccccccc}
\toprule
& \multicolumn{3}{c}{5-way} & \multicolumn{3}{c}{4-way} & \multicolumn{3}{c}{3-way} & \multicolumn{3}{c}{2-way} \\\cmidrule(lr){2-4} \cmidrule(lr){5-7} \cmidrule(lr){8-10} \cmidrule(lr){11-13}
Model & P & R & F & P & R & F & P & R & F & P & R & F \\ \midrule
\texttt{XGBoost} & 38.17 & 40.06 & 34.59 & 46.16 & 44.12 & 41.29 & 56.19 & 54.64 & 51.98 & 64.5 & 79.56 & 63.67 \\\midrule
\texttt{c-RNN} & 52.42 & 48.49 & 49.62 & 54.60 & 54.65 & 54.19 & 56.78 & 53.50 & 54.68 & 67.66 & 62.89 & 64.59 \\
\texttt{ +SGNS} &54.54 & 51.27 & 51.56 & \underline{57.78} & \underline{56.75} & \underline{57.16} & \underline{59.57} & \underline{56.19} & \underline{57.46} & \underline{67.74} & \underline{65.33} & \underline{66.37} \\
\texttt{ +BERT} & \textbf{64.99} & \textbf{63.60} & \textbf{63.85} & \textbf{68.22} & \textbf{69.48} & \textbf{68.17} & \textbf{70.36} & \textbf{68.60} & \textbf{69.13} & \textbf{78.35} & \textbf{73.51} & \textbf{75.59} \\
&&& (+28.68\%) &&& (+25.80\%) &&& (+26.43\%) &&& (+17.03\%) \\
\texttt{ConATT} & 51.78 & 48.28 & 49.25 & 54.27 & 53.08 & 52.98 & 53.67 & 49.47 & 50.79 & 63.25 & 56.92 & 58.28 \\
\texttt{ +SGNS} & \underline{55.44} & \underline{52.13} & \underline{53.09} & 55.88 & 54.94 & 54.18 & 55.01 & 53.06 & 53.87 & 64.98 & 61.38 & 62.69 \\
\bottomrule
\end{tabular}}
\caption{Evaluation results of the Chinese RFS systems on \ontonoteszh.}
\label{tab:zh_cls_result}
\end{table*}

\section{Experiments}

In what follows, we first provide an overview of the implementation details of the RFS models. To understand what linguistic information can be learnt by each model, we introduce a series of probing experiments.
We then discuss the performance of these models and answer the hypotheses.

\begin{table*}[t!]
\small
\centering
\begin{tabular}{llccccccc}
\toprule
Model & Type & DisStat & SenStat & Syn & DistAnt & IntRef & LocPro & GloPro \\ \midrule
\texttt{Random} & - & \makecell[c]{49.99\\(49.77)} & \makecell[c]{33.06\\(32.27)} & \makecell[c]{50.10\\(50.10)} & \makecell[c]{25.17\\(23.75)} & \makecell[c]{33.09\\(32.40)} & \makecell[c]{49.94\\(48.21)} & \makecell[c]{50.38\\(49.53)} \\
\texttt{Majority} & - & \makecell[c]{55.95\\(35.88)} & \makecell[c]{44.05\\(20.39)} & \makecell[c]{50.14\\(33.39)} & \makecell[c]{44.05\\(15.29)} & \makecell[c]{44.05\\(20.39)}  & \makecell[c]{68.08\\(40.50)} & \makecell[c]{63.08\\(38.68)} \\ \midrule
\multirow{3}{*}{\texttt{c-RNN}} 
& 4-way & \makecell[c]{64.73\\(63.39)} & \makecell[c]{54.41\\(50.76)} & \makecell[c]{74.73\\(74.67)} & \makecell[c]{51.66\\(36.31)} & \makecell[c]{50.52\\(44.81)} & \makecell[c]{74.57\\(67.86)} & \makecell[c]{63.89\\(50.32)}  \\
& 3-way & \makecell[c]{64.24\\(63.30)} & \makecell[c]{53.94\\(50.45)} & \makecell[c]{75.57\\(75.55)} & \makecell[c]{52.02\\(36.78)} & \makecell[c]{49.76\\(42.83)} & \makecell[c]{74.96\\(68.26)} & \makecell[c]{64.00\\(49.71)}  \\
& 2-way & \makecell[c]{64.45\\(63.31)} & \makecell[c]{53.55\\(49.72)} & \makecell[c]{73.90\\(73.82)} & \makecell[c]{51.55\\(35.75)} & \makecell[c]{49.67\\(43.03)} & \makecell[c]{73.50\\(65.72)} & \makecell[c]{63.39\\(45.76)}  \\ \midrule
\multirow{3}{*}{\makecell[l]{\texttt{c-RNN}\\\texttt{+GloVe}}} 
& 4-way & \makecell[c]{65.00\\(64.24)} & \makecell[c]{54.40\\(51.39)} & \makecell[c]{76.75\\(76.75)} & \makecell[c]{51.95\\(37.09)} & \makecell[c]{50.65\\(44.94)} & \makecell[c]{74.25\\(67.26)} & \makecell[c]{64.14\\(51.44)}  \\
& 3-way & \makecell[c]{65.17\\(64.44)} & \makecell[c]{55.14\\(52.69)} & \makecell[c]{78.06\\(78.06)} & \makecell[c]{52.81\\(37.55)} & \makecell[c]{50.73\\(45.89)} & \makecell[c]{75.46\\(70.66)} & \makecell[c]{64.67\\(53.28)}  \\
& 2-way & \makecell[c]{65.07\\(64.26)} & \makecell[c]{53.55\\(49.34)} & \makecell[c]{75.22\\(75.06)} & \makecell[c]{51.20\\(35.87)} & \makecell[c]{50.78\\(45.04)} & \makecell[c]{73.91\\(67.22)} & \makecell[c]{63.26\\(47.49)} \\ \midrule
\multirow{3}{*}{\makecell[l]{\texttt{c-RNN}\\\texttt{+BERT}}} 
& 4-way & \makecell[c]{86.00\\(85.67)} & \makecell[c]{72.17\\(69.46)} & \makecell[c]{79.83\\(79.73)} & \makecell[c]{66.53\\(50.36)} & \makecell[c]{69.85\\(65.99)} & \makecell[c]{82.32\\(80.08)} & \makecell[c]{68.47\\(60.06)}  \\
& 3-way & \makecell[c]{83.74\\(83.42)} & \makecell[c]{71.56\\(68.90)} & \makecell[c]{81.17\\(81.15)} & \makecell[c]{65.35\\(49.10)} & \makecell[c]{68.03\\(63.62)} & \makecell[c]{85.05\\(82.38)} & \makecell[c]{67.82\\(61.93)}  \\
& 2-way & \makecell[c]{81.82\\(81.12)} & \makecell[c]{69.33\\(67.07)} & \makecell[c]{78.05\\(77.89)} & \makecell[c]{63.46\\(47.97)} & \makecell[c]{65.11\\(62.06)} & \makecell[c]{81.85\\(77.45)} & \makecell[c]{66.35\\(53.37)} \\ \midrule
\multirow{3}{*}{\texttt{ConATT}} 
& 4-way & \makecell[c]{64.37\\(62.95)} & \makecell[c]{52.20\\(46.63)} & \makecell[c]{73.37\\(73.34)} & \makecell[c]{49.74\\(33.33)} & \makecell[c]{49.55\\(43.52)} & \makecell[c]{74.04\\(66.30)} & \makecell[c]{63.57\\(48.89)} \\
& 3-way & \makecell[c]{64.28\\(61.87)} & \makecell[c]{51.96\\(45.92)} & \makecell[c]{74.79\\(74.76)} & \makecell[c]{49.25\\(31.91)} & \makecell[c]{49.21\\(41.64)} & \makecell[c]{73.89\\(67.51)} & \makecell[c]{63.25\\(48.61)} \\
& 2-way & \makecell[c]{62.07\\(59.46)} & \makecell[c]{49.45\\(41.73)} & \makecell[c]{64.44\\(63.72)} & \makecell[c]{48.05\\(30.18)} & \makecell[c]{47.85\\(40.85)} & \makecell[c]{71.24\\(59.96)} & \makecell[c]{63.32\\(47.51)} \\ \midrule
\multirow{3}{*}{\makecell[l]{\texttt{ConATT}\\\texttt{+GloVe}}} 
& 4-way & \makecell[c]{65.39\\(63.41)} & \makecell[c]{53.51\\(50.49)} & \makecell[c]{79.96\\(79.95)} & \makecell[c]{51.51\\(36.03)} & \makecell[c]{50.52\\(43.17)} & \makecell[c]{76.05\\(70.27)} & \makecell[c]{63.79\\(49.86)} \\
& 3-way & \makecell[c]{63.72\\(61.79)} & \makecell[c]{52.13\\(45.39)} & \makecell[c]{79.03\\(79.00)} & \makecell[c]{49.48\\(33.03)} & \makecell[c]{49.43\\(41.53)} & \makecell[c]{74.86\\(68.46)} & \makecell[c]{63.31\\(48.97)} \\
& 2-way & \makecell[c]{63.77\\(61.56)} & \makecell[c]{50.73\\(44.35)} & \makecell[c]{74.20\\(73.97)} & \makecell[c]{48.77\\(31.53)} & \makecell[c]{49.24\\(42.81)} & \makecell[c]{72.31\\(63.31)} & \makecell[c]{63.15\\(48.39)} \\
\bottomrule
\end{tabular}
\caption{Results of the English RFS models on each probing task on the \ontonotesen dataset. A in A(B) is the accuracy and B is the macro F1.}
\label{tab:en_prob}
\end{table*}
\begin{table*}[t!]
\small
\centering
\begin{tabular}{llccccccc}
\toprule
Model & Type & DisStat & SenStat & Syn & DistAnt & IntRef & LocPro & GloPro \\ \midrule
\texttt{Random} & - & \makecell[c]{50.20\\(49.93)} & \makecell[c]{33.18\\(32.70)} & \makecell[c]{50.11\\(49.79)} & \makecell[c]{25.02\\(23.81)} & \makecell[c]{33.56\\(33.01)} & \makecell[c]{50.12\\(46.44)} & \makecell[c]{50.00\\(44.27)} \\
\texttt{Majority} & - & \makecell[c]{57.30\\(36.43)} & \makecell[c]{42.70\\(19.95)} & \makecell[c]{57.79\\(36.62)} & \makecell[c]{42.70\\(14.96)} & \makecell[c]{42.70\\(19.95)}  & \makecell[c]{76.27\\(43.27)} & \makecell[c]{81.13\\(45.09)} \\ \midrule
\multirow{3}{*}{\texttt{c-RNN}} 
& 5-way & \makecell[c]{65.14\\(62.80)} & \makecell[c]{48.85\\(45.89)} & \makecell[c]{76.79\\(75.94)} & \makecell[c]{46.50\\(28.49)} & \makecell[c]{48.72\\(45.78)} & \makecell[c]{79.12\\(65.54)} & \makecell[c]{82.57\\(52.03)}  \\
& 4-way & \makecell[c]{64.60\\(61.80)} & \makecell[c]{48.76\\(43.39)} & \makecell[c]{76.30\\(74.74)} & \makecell[c]{45.75\\(27.73)} & \makecell[c]{47.84\\(44.65)} & \makecell[c]{79.11\\(63.44)} & \makecell[c]{81.97\\(46.64)}  \\
& 3-way & \makecell[c]{63.55\\(61.19)} & \makecell[c]{47.52\\(41.52)} & \makecell[c]{77.13\\(76.11)} & \makecell[c]{45.69\\(26.43)} & \makecell[c]{46.60\\(41.13)} & \makecell[c]{78.11\\(61.70)} & \makecell[c]{82.02\\(45.76)}  \\
& 2-way & \makecell[c]{61.32\\(58.06)} & \makecell[c]{46.09\\(36.30)} & \makecell[c]{77.95\\(76.96)} & \makecell[c]{45.23\\(24.11)} & \makecell[c]{45.71\\(36.49)} & \makecell[c]{77.86\\(58.82)} & \makecell[c]{82.11\\(45.54)}  \\ \midrule
\multirow{3}{*}{\makecell[l]{\texttt{c-RNN}\\\texttt{+SGNS}}} 
& 5-way & \makecell[c]{65.75\\(63.52)} & \makecell[c]{50.24\\(47.24)} & \makecell[c]{78.36\\(77.28)} & \makecell[c]{47.48\\(30.71)} & \makecell[c]{49.66\\(46.13)} & \makecell[c]{79.33\\(66.11)} & \makecell[c]{82.21\\(50.37)}  \\
& 4-way & \makecell[c]{66.07\\(62.90)} & \makecell[c]{50.93\\(46.96)} & \makecell[c]{78.41\\(77.18)} & \makecell[c]{47.64\\(30.78)} & \makecell[c]{50.57\\(47.81)} & \makecell[c]{80.11\\(66.16)} & \makecell[c]{82.24\\(48.20)}  \\
& 3-way & \makecell[c]{64.70\\(62.87)} & \makecell[c]{48.24\\(42.54)} & \makecell[c]{79.02\\(77.81)} & \makecell[c]{46.27\\(27.51)} & \makecell[c]{47.48\\(43.59)} & \makecell[c]{79.35\\(64.17)} & \makecell[c]{82.01\\(46.11)}  \\
& 2-way & \makecell[c]{62.48\\(60.45)} & \makecell[c]{46.30\\(38.24)} & \makecell[c]{78.50\\(77.12)} & \makecell[c]{45.38\\(24.27)} & \makecell[c]{44.82\\(37.61)} & \makecell[c]{77.72\\(64.09)} & \makecell[c]{81.93\\(46.12)}  \\ \midrule
\multirow{3}{*}{\makecell[l]{\texttt{c-RNN}\\\texttt{+BERT}}}
& 5-way & \makecell[c]{76.17\\(75.20)} & \makecell[c]{59.58\\(57.07)} & \makecell[c]{79.42\\(78.68)} & \makecell[c]{56.14\\(39.54)} & \makecell[c]{59.89\\(57.69)} & \makecell[c]{81.86\\(70.93)} & \makecell[c]{82.05\\(55.17)}  \\
& 4-way & \makecell[c]{75.32\\(73.96)} & \makecell[c]{59.69\\(57.66)} & \makecell[c]{78.86\\(78.15)} & \makecell[c]{56.66\\(37.12)} & \makecell[c]{60.27\\(56.90)} & \makecell[c]{81.95\\(69.68)} & \makecell[c]{81.96\\(46.60)} \\
& 3-way & \makecell[c]{74.46\\(73.77)} & \makecell[c]{58.41\\(56.29)} & \makecell[c]{80.48\\(79.67)} & \makecell[c]{55.91\\(35.77)} & \makecell[c]{59.39\\(55.96)} & \makecell[c]{82.71\\(73.24)} & \makecell[c]{81.91\\(45.59)} \\
& 2-way & \makecell[c]{69.20\\(68.10)} & \makecell[c]{55.16\\(52.08)} & \makecell[c]{80.68\\(79.84)} & \makecell[c]{51.74\\(29.71)} & \makecell[c]{51.73\\(52.36)} & \makecell[c]{81.43\\(71.30)} & \makecell[c]{82.05\\(45.07)} \\ \midrule
\multirow{3}{*}{\texttt{ConATT}} 
& 5-way & \makecell[c]{65.36\\(62.33)} & \makecell[c]{48.50\\(43.17)} & \makecell[c]{75.14\\(73.94)} & \makecell[c]{46.44\\(28.92)} & \makecell[c]{48.25\\(45.05)} & \makecell[c]{78.90\\(63.99)} & \makecell[c]{82.02\\(47,16)}  \\
& 4-way & \makecell[c]{65.07\\(61.91)} & \makecell[c]{48.40\\(43.15)} & \makecell[c]{70.38\\(67.48)} & \makecell[c]{45.95\\(26.41)} & \makecell[c]{48.16\\(44.15)} & \makecell[c]{77.89\\(57.31)} & \makecell[c]{82.22\\(47.27)} \\
& 3-way & \makecell[c]{62.93\\(59.54)} & \makecell[c]{45.14\\(39.55)} & \makecell[c]{70.38\\(68.78)} & \makecell[c]{43.85\\(24.47)} & \makecell[c]{45.28\\(39.13)} & \makecell[c]{77.34\\(55.27)} & \makecell[c]{82.06\\(45.73)} \\
& 2-way & \makecell[c]{60.55\\(52.10)} & \makecell[c]{44.21\\(32.85)} & \makecell[c]{68.33\\(65.67)} & \makecell[c]{43.75\\(21.78)} & \makecell[c]{44.36\\(32.66)} & \makecell[c]{76.37\\(49.38)} & \makecell[c]{82.07\\(45.35)} \\ \midrule
\multirow{3}{*}{\makecell[l]{\texttt{ConATT}\\\texttt{+SGNS}}} 
& 5-way & \makecell[c]{66.65\\(63.48)} & \makecell[c]{49.57\\(45.60)} & \makecell[c]{78.18\\(77.27)} & \makecell[c]{46.34\\(29.77)} & \makecell[c]{49.76\\(46.84)} & \makecell[c]{79.35\\(64.16)} & \makecell[c]{81.65\\(50.72)}  \\
& 4-way & \makecell[c]{66.09\\(61.97)} & \makecell[c]{49.43\\(44.63)} & \makecell[c]{75.87\\(74.65)} & \makecell[c]{46.04\\(28.19)} & \makecell[c]{49.20\\(46.61)} & \makecell[c]{79.50\\(64.49)} & \makecell[c]{82.22\\(47.27)} \\
& 3-way & \makecell[c]{62.84\\(58.79)} & \makecell[c]{46.51\\(38.78)} & \makecell[c]{75.15\\(74.09)} & \makecell[c]{44.99\\(24.66)} & \makecell[c]{45.76\\(38.51)} & \makecell[c]{78.12\\(60.19)} & \makecell[c]{82.06\\(45.73)} \\
& 2-way & \makecell[c]{62.65\\(60.09)} & \makecell[c]{46.76\\(39.53)} & \makecell[c]{74.17\\(72.90)} & \makecell[c]{44.31\\(22.13)} & \makecell[c]{44.84\\(34.88)} & \makecell[c]{77.53\\(61.43)} & \makecell[c]{82.07\\(45.35)} \\
\bottomrule
\end{tabular}
\caption{Results of the Chinese RFS models on each probing task on the \ontonoteszh.}
\label{tab:zh_prob}
\end{table*}

\subsection{Baseline and Implementation Details}

Following~\citet{chen-etal-2021-neural-referential}, we used a feature-based model, \texttt{XGBoost}~\citep{chen2015xgboost}, as our baseline. 
For pre-trained word embeddings, we used \texttt{Glove}~\citep{pennington-etal-2014-glove} for English and \texttt{SGNS}~\citep{li-etal-2018-analogical} for Chinese; for \bert, we used ``bert-base-cased'' for English and ``bert-base-chinese'' for Chinese.\footnote{(1) English Glove: \url{https://nlp.stanford.edu/projects/glove/}; (2) Chinese SGNS: \url{https://github.com/Embedding/Chinese-Word-Vectors}; (3) English \bert: \url{huggingface.co/bert-base-cased}; and (4) Chinese \bert: \url{https://huggingface.co/bert-base-chinese}.}
Since Chinese \bert is a character-based model, we use all Chinese models character-based. The results of the word-based models 
can be found in Appendix~\ref{sec:app_word}.

We tuned the hyper-parameters of each of our neural models on the development set and chose the setting with the best macro F1 score. For training, we used a single Tesla V100. For the baseline \texttt{XGBoost} models, we set the learning rate to 0.05, the minimum split loss to 0.01, the maximum depth of a tree to 5, and the sub-sample ratio of the training instances to 0.5.
We report macro-averaged precision, recall, and F1 on the test set. We run each model 5x and report the average performance.

\subsection{Probing RFS Models}

To test the hypotheses in \S\ref{sec:hypo} (especially $\mathcal{H}_3$), we probed each RFS model using probing classifiers. Specifically, after training an RFS model, we extracted its hidden representations and used them to train a probing classifier for a particular linguistic feature. The performance of the probing classifier indicates how well the RFS model learns the feature~\citep{belinkov-etal-2017-neural, giulianelli-etal-2018-hood}. 

\paragraph{Probing Tasks.}

We used the probing tasks defined in \citet{chen-etal-2021-neural-referential}. These tasks pertain to four classes of features, namely referential status (DisStat and SenStat), syntactic position (Syn), recency (DistAnt and IntRef), and discourse structure prominence (LocPro, GloPro). These features have been shown to matter for RFS in linguistic literature \citep{ariel1990accessing,gundel1993cognitive,arnold2010speakers,von2019discourse}. The definition of each probing task is as follows: 
(1) \textbf{DisStat}: This feature has 2 values: (a) \texttt{discourse-old} (the entity appeared in the previous context), and (b) \term{discourse-new} (it did not);
(2) \textbf{SenStat}: The sentence-level referential status feature has 3 values: (a) \term{sentence-new} (the RE is the first mention of the entity in the sentence), (b) \term{sentence-old} (the RE is not the first mention of the entity in the sentence), and (c) \term{first-mention} (the RE is the first mention of the entity in the discourse);
(3) \textbf{Syn}: The syntax probing task is a binary classification task with values (a) \term{subject} and (b) \term{object};
(4) \textbf{DistAnt}: It contains four values: the entity and its antecedent are (a) in \term{same} sentence, (b) \term{one} sentence apart, (c) \term{more than one} sentence apart, and (d) the entity is a \term{first-mention} (to distinguish first mentions from subsequent mentions); 
(5) \textbf{IntRef}: This feature asks whether there is an intervening referent between the target RE and its nearest antecedent. There are 3 possible values: (a) the target entity is a \term{first-mention}, (b) the previous RE refers to the \term{same entity}, and (c) the previous RE refers to a \term{different entity};
(6) \textbf{LocPro}: is a hybrid of DisStat and Syn. It has 2 values: (a) \term{locally prominent}, and (b) \term{not locally prominent}. An entity is said to be locally prominent if it is both ``discourse- old" and ``realised as a subject";
(7) \textbf{GloPro}: This is a binary feature with two possible values: (a) globally prominent, and (b) not globally prominent. The most frequent entity in a text is marked as globally prominent.

\paragraph{Probing Classifiers.}

Following~\citet{chen-etal-2021-neural-referential}, we use a logistic regression classifier as our probing classifier.
When probing, we use $R$ (see Equation~\ref{eq:r_1} and~\ref{eq:r_2}) of the models with the best RFS performance on the development set as input representations.
We evaluate probing classifiers using the accuracy and macro-averaged F1 scores. 
We run each probing classifier 5 times and report the averaged value. 
We use 2 baselines: (1) \texttt{random}: it randomly assigns a label to each input; and (2) \texttt{majority}: it assigns the most frequent label in the given probing task to the inputs. 



\begin{figure*}[t]
    \centering
    \includegraphics[scale=0.25]{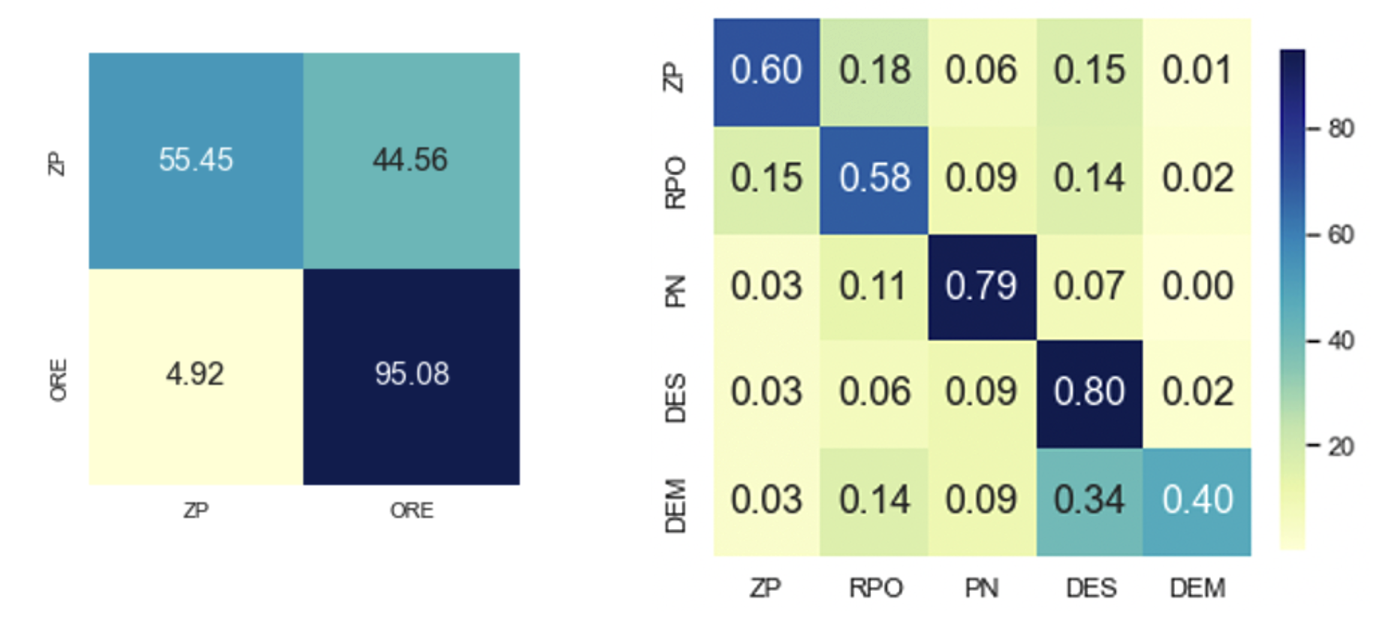}
    \caption{Confusion Matrix for Chinese 2-way \crnn+\bert (left) and 5-way \crnn+\bert (right) where ORE is overt RE, PRO is pronoun, PN is proper name, DES is description, and DEM is demonstrative.}
    \label{fig:cm}
\end{figure*}

\begin{table*}[t]
\centering
\begin{tabular}{lccccccccc}
\toprule
 & \multicolumn{3}{c}{4-way} & \multicolumn{3}{c}{3-way} & \multicolumn{3}{c}{2-way} \\\cmidrule(lr){2-4} \cmidrule(lr){5-7} \cmidrule(lr){8-10}
Model & P & R & F & P & R & F & P & R & F  \\ \midrule
\texttt{c-RNN}  & 50.77 & 45.89 & 46.38 & \underline{60.83} & 59.56 & \underline{59.94} & 73.33 & \underline{72.58} & \underline{72.84} \\
\texttt{ +Glove} & \underline{53.47} & \textbf{49.49} & \textbf{50.44} & \textbf{61.72} & \textbf{60.66} & \textbf{60.98} & \textbf{75.06} & \textbf{73.96} & \textbf{74.32} \\
\texttt{ConATT} & 52.32 & 45.88 & 46.89 & 59.66 & 58.71 & 59.08 & 71.86 & 71.38 & 71.56 \\
\texttt{ +Glove} & \textbf{54.55} & \underline{47.56} & \underline{48.14} & 59.75 & \underline{60.05} & 59.85 & \underline{73.84} & 72.32 & 72.66 \\
\bottomrule
\end{tabular}
\caption{Evaluation results of RFS systems on \ontonotesen with entity tags.}
\label{tab:en_cls_result_underscore_intext}
\end{table*}

\subsection{Experimental Results} \label{sec:results}

\paragraph{Results on each Language.} 

Table~\ref{tab:en_cls_result} and~\ref{tab:zh_cls_result} show the results of each model on \ontonotesen and \ontonoteszh. In both languages, all neural RFS models defeat the baseline in 4-way and 5-way classifications, while models that does not use \bert have on-par or worse performance in 3-way and 2-way classifications. This suggests that feature-based models with linguistically-informed features can build remarkably good systems for RFS, but their performance decreases dramatically as the task becomes more fine-grained.

As for $\mathcal{H}_1$, word embeddings always improve RFS performance. 
The RFS tasks in both languages benefit strongly from using \bert. For instance, if we compare \crnn+\bert to \crnn for the full RFS tasks (i.e., 5-way classification in Chinese and 4-way classification in English), \crnn+\bert improves the performance (F1 score) from 62.38 to 74.59 in English and from 49.62 to 63.85 in Chinese.

In both languages, contrary to our expectation $\mathcal{H}_2$, \conatt performs worse than \crnn. Probing results presented in Tables~\ref{tab:en_prob} and~\ref{tab:zh_prob} provide some explanations:
in English, \conatt learns less information about referential status, syntactic position, and recency than \crnn, and in Chinese, \conatt performs significantly worse than \crnn in acquiring information about syntactic position. 

Meanwhile, the results of the probing experiments suggest
that expectation $\mathcal{H}_3$, that models that learn more useful information perform better, is true. Further evidence is provided by the observations that (1) \bert defeats all other models in almost all probing tasks 
and, therefore, defeats all other models by a large margin;
and (2) pre-trained word embeddings (\texttt{GloVe} and \texttt{SGNS}) help each model learn significantly more information about almost every feature except GloPro, and, therefore, improve RFS performance.

\paragraph{English vs. Chinese.} 

In line with our expectation $\mathcal{H}_4$, models that work well for English also work well on modelling ZP in Chinese. However, deciding whether to use a ZP or an overt RE is generally harder than pronominalisation. For example, \crnn achieves an F-score of 75.7 for the English 2-way task, while it is only 64.6 for Chinese. 

Figure~\ref{fig:cm} shows the confusion matrices for the Chinese \crnn+\bert 2-way and 5-way classifications. 
By comparing them, we find that fine-grained supervision helps with the choice between ZPs and overt REs.
Focusing on 5-way classification, ZPs are quite often confused with pronouns. Linguistic theory suggests that attenuated forms such as pronouns and ZPs 
happen when the target referent is salient enough \citep{ariel2001accessibility}. It is understandable that ZPs and pronouns are confused because it is hard for a model to make such a fine-grained decision about when the target referent is salient enough for pronominalisation but not for pro-drop.

The results of both Chinese and English RFS tasks improve dramatically when using the contextual language model \bert. This is consistent with the probing results: in both languages, \bert helps a lot in acquiring all linguistic information except GloPro. To test our last hypothesis $\mathcal{H}_5$, we compute how much \crnn gains from using \bert compared to not using \bert and report the numbers in Table~\ref{tab:en_cls_result} and~\ref{tab:zh_cls_result}. On average, \crnn gains 17.60\% from using \bert in English and 24.48\% in Chinese. The results suggest that Chinese RFS benefits more from using \bert than English RFS. Nevertheless, we still cannot make conclusive statements about $\mathcal{H}_5$. Strictly speaking, these percentages are not directly comparable
and the comparison cannot be fully controlled because for example: (1) the data is not fully parallel, and (2) the RFS tasks defined for the two languages differ from each other. For instance, unlike  English RFS, Chinese RFS considers an extra category, namely ZP.


\paragraph{Lexical Tags vs. Entity Tags.}

To chart the benefits of lexical tags, we also ran models of~\citet{chen-etal-2021-neural-referential} on a version of \ontonotesen, in which entity tags are used instead of lexical tags. The results are presented in Table~\ref{tab:en_cls_result_underscore_intext}.
Comparing this table to Table~\ref{tab:en_cls_result}, we see that the performance of each model decreases significantly when the entity tags are used, especially in the 4-way and 3-way classifications.
For example, the F-score of the 4-way \crnn+\texttt{Glove} model decreases from 64.56 to 50.44. As expected, these tags prevent the models from handling unseen entities. 

\section{Conclusion}

To address the problem that all previous assessments of neural REG/RFS models were only tested on \webnlg, 
we built a realistic multilingual (English and Chinese) dataset based on the \ontonotes dataset, modified the RFS models accordingly and assessed them on this dataset. Although a few outcomes were against our expectations (e.g. \conatt performed worse than \crnn), we found that our results are explainable using probing experiments. 
For example, models that use \bert, which performs best in the probing experiments, also beats all other models in RFS.

We also compared the English RFS to the Chinese RFS, which 
uses ZPs frequently and depends more on context than English. We found that RFS models that work for English can also model Chinese ZPs. In line with the idea that Chinese relies more on context than English, the results suggest that Chinese RFS models benefited more from using contextualised language model \bert than those of English. However, as discussed, this needs to be further verified with more controlled experiments.


In future, we plan to extend our work from the following three perspectives: (1) testing other model explanation techniques, e.g., probing classifiers with control tasks~\citep{hewitt-liang-2019-designing} and attention analysis~\citep{bibal-etal-2022-attention}; (2) assessing and probing RFS models on other languages (such as languages that are morphologically rich
); and (3) trying more probing tasks based on factors that influence RFS, such as animacy, competition and positional attributes (see \citet{same-van-deemter-2020-linguistic} for more details).

\bibliography{anthology,custom}
\bibliographystyle{acl_natbib}

\appendix

\newpage

\section{Results on the Whole \ontonoteszh Dataset} \label{sec:app_whole} 

\begin{table*}
\small
\centering
\begin{tabular}{lcccccccccccc}
\toprule
& \multicolumn{3}{c}{5-way} & \multicolumn{3}{c}{4-way} & \multicolumn{3}{c}{3-way} & \multicolumn{3}{c}{2-way} \\\cmidrule(lr){2-4} \cmidrule(lr){5-7} \cmidrule(lr){8-10} \cmidrule(lr){11-13}
Model & P & R & F & P & R & F & P & R & F & P & R & F \\ \midrule
\texttt{c-RNN} & 52.36 & 47.91 & 48.97 & 54.14 & 52.40 & 53.06 & 55.30 & 52.99 & 53.86 & 64.88 & 62.81 & 63.68 \\
\texttt{ +SGNS} & 56.67 & 53.82 & 54.30 & 59.38 & 57.40 & 58.23 & 59.58 & 56.66 & 57.78 & 67.75 & 66.28 & 66.91 \\
\texttt{ConATT} & 50.41 & 45.45 & 46.86 & 51.27 & 49.80 & 50.35 & 59.06 & 54.43 & 56.11 & 63.71 & 63.75 & 63.73 \\
\texttt{ +SGNS} & 52.33 & 48.60 & 49.37 & 53.48 & 51.64 & 52.38 & 60.53 & 56.18 & 57.69 & 67.86 & 64.97 & 65.95 \\
\bottomrule
\end{tabular}
\caption{Evaluation results of our word-based Chinese RFS systems on the whole \ontonoteszh dataset.}
\label{tab:zh_whole_cls_result}
\end{table*}
\begin{table*}
\small
\centering
\begin{tabular}{lcccccccccccc}
\toprule
& \multicolumn{3}{c}{5-way} & \multicolumn{3}{c}{4-way} & \multicolumn{3}{c}{3-way} & \multicolumn{3}{c}{2-way} \\\cmidrule(lr){2-4} \cmidrule(lr){5-7} \cmidrule(lr){8-10} \cmidrule(lr){11-13}
Model & P & R & F & P & R & F & P & R & F & P & R & F \\ \midrule
\texttt{c-RNN} & 51.13 & 47.14 & 48.63 & 54.70 & 54.02 & 54.18 & 57.63 & 53.79 & 55.16 & 66.19 & 63.22 & 64.40 \\
\texttt{ +SGNS} & 53.40 & 53.33 & 53.16 & 57.91 & 59.12 & 58.19 & 60.17 & 57.49 & 58.52 & 70.87 & 65.22 & 67.30 \\
\texttt{ConATT} & 48.52 & 45.15 & 46.26 & 56.34 & 49.92 & 49.26 & 56.24 & 55.70 & 55.94 & 65.33 & 64.28 & 64.75 \\
\texttt{ +SGNS} & 50.58 & 47.04 & 48.31 & 54.68 & 51.85 & 52.62 & 59.93 & 55.79 & 57.32 & 67.15 & 65.29 & 66.11 \\
\bottomrule
\end{tabular}
\caption{Evaluation results of our word-based Chinese RFS systems on a subset of the original \ontonoteszh dataset, each text of which contains less than 512 characters.}
\label{tab:zh_w_cls_result}
\end{table*}

The Chinese experiments in this paper were conducted on a subset of the original \ontonotes, each text of which contains less than 512 characters, since Chinese \bert can only accept texts shorter than 512 characters. 
For reference, we also tested models other than \bert on the whole \ontonoteszh dataset. 
In the whole \ontonoteszh dataset, there are 73607, 10008, and 12096 samples in the training, development, and test sets, respectively.
Table \ref{tab:zh_whole_cls_result} shows the results of the word-based Chinese RFS models on the whole \ontonoteszh dataset.

Comparing Table~\ref{tab:zh_whole_cls_result} with Table~\ref{tab:zh_w_cls_result}, we observe that the results are quite similar. The only exception is that the performance of \crnn decreases from 55.16 to 53.86 in the 3-way classification, while the performance of \conatt does not change much.

\section{Results of Using Word-based Models on \ontonoteszh} \label{sec:app_word}

To conduct a fair comparison between \bert and other models, we built all our Chinese RFS models character-based. 
To justify this decision, we also test word-based models on \ontonoteszh. Table \ref{tab:zh_cls_result} shows the results of the word-based Chinese models.

Comparing the results in Table~\ref{tab:zh_w_cls_result} and Table~\ref{tab:zh_cls_result}, there are slight differences, but these differences do not change our conclusions. For example, all models still perform worse than \crnn+\bert by a large margin. \conatt can slightly defeat \crnn in the 3-way and 2-way classifications but performs significantly worse in other settings.



\end{document}